\pdfoutput=1

\documentclass[11pt]{article}

\usepackage[preprint]{acl}

\usepackage{times}
\usepackage{latexsym}
\usepackage{booktabs}
\usepackage{multirow,multicol}
\usepackage{url}

\usepackage[T1]{fontenc}

\usepackage[utf8]{inputenc}

\usepackage{microtype}

\usepackage{inconsolata}
\usepackage{linguex}
\usepackage{graphicx}
\usepackage{xcolor}

\title{A Dataset for Misogyny Detection in Polish Social Media: Transferability of Hateful Speech Detection from General to Specific Types of Hate}

\author{Ewelina Gajewska \\
  Warsaw University of Technology \\
  Warsaw, Poland \\
  \texttt{ewelina.gajewska.dokt@pw.edu.pl}}

\begin{document}
\maketitle
\begin{abstract}
Harmful language detection is commonly evaluated on broadly defined hate speech and toxicity benchmarks, yet it remains unclear whether models that successfully detect harmful language in general can transfer this capability to more specific forms of hate. We investigate this problem for misogyny detection in Polish, an area for which dedicated resources remain scarce. We introduce a new dataset of Polish social media comments annotated for misogyny, gender-directed hateful speech. Using this resource, we first assess the reliability of LLM-assisted annotation and then evaluate general-to-specific transfer by comparing models trained on general harmful language data, safety-specialized models, and models fine-tuned directly for misogyny detection. Our results show that strong performance on general harmful language detection does not transfer uniformly to misogyny; safety-specialized models demonstrate substantially stronger generalization ability. The study contributes a new Polish resource for gender-directed harmful language and demonstrates that the scope and granularity of prior safety training are important factors in transferring harmful language detection from general to specific forms of hate.

\footnotesize{\textcolor{red}{Warning}: This article contains examples of hateful language.}

\end{abstract}

\section{Introduction}
Social media platforms have become central venues for public debate, interpersonal communication, and opinion formation. While these platforms facilitate participation in public discourse, they also provide fertile ground for various forms of abusive and discriminatory language. Among these, misogynistic discourse remains a persistent problem, affecting women across diverse social, professional, and public contexts. Expressions of hostility toward women range from explicit insults and derogatory statements to more subtle forms of prejudice embedded in stereotypes, mockery, and gender-based assumptions. The scale and velocity of online communication make manual moderation increasingly impractical, creating a pressing need for reliable automated systems capable of identifying harmful content.

The development of such systems relies heavily on the availability of high-quality annotated datasets. Although hate speech detection has become an established area of Natural Language Processing (NLP), research resources remain unevenly distributed across languages. Compared to English and several other high-resource languages, Polish continues to be underrepresented in the hate speech detection literature. Existing Polish-language datasets have primarily focused on broader forms of online toxicity, political hostility, offensive language, or hate speech directed at ethnic, national, religious, or social groups. In contrast, misogyny has received considerably less attention, despite its prevalence in online discussions. Even more scarce are datasets that capture implicit forms of misogyny, where hostility is expressed indirectly through stereotypes, insinuations, or socially normalized forms of gender prejudice rather than overt insults.

This gap is particularly important because implicit misogyny presents a substantial challenge for contemporary NLP systems \citep{Gajewska2026ImplicitHateSpeech,samghabadi2020aggression}. While explicit hate speech contains lexical cues that facilitate automatic identification, stereotype-based or indirectly expressed hostility frequently requires contextual and pragmatic interpretation \citep{ocampo2023depth,elsherief2021latent}. As a result, models trained on existing hate speech datasets may fail to recognize harmful content that does not conform to prototypical patterns of hateful language. 
More broadly, this raises the question of whether harmful language detection capabilities learned from general-purpose datasets transfer to more specific forms of discriminatory language. Understanding this general-to-specific transfer is important for determining what existing moderation benchmarks actually tell us about model performance on particular manifestations of online harm. 

The present study addresses three limitations in the current landscape of Polish harmful language detection. First, existing datasets provide limited coverage of gender-directed abuse, leaving misogyny and stereotype-based hostility relatively underexplored. Second, little is known about whether models trained to detect harmful language in general can transfer this capability to more specific forms of hate, or whether broader safety-oriented specialization provides more transferable representations. Third, the capabilities of different model families for Polish misogyny detection remain insufficiently understood, particularly when distinguishing misogyny from subtler gender stereotyping. These limitations constrain both the development of target-specific moderation systems and our understanding of what general harmful language competence actually transfers across domains and types of harm.

To address these challenges, we construct a new dataset of Polish social media comments collected from discussions surrounding a highly visible media controversy involving a prominent female public figure. The resulting corpus is manually annotated according to three categories: misogyny, gender stereotyping, and neutral discourse. The annotation process combines the judgments of two frontier large language models (LLMs) with expert adjudication in cases of disagreement. Based on this dataset, we investigate 3 research questions:
\begin{itemize}
    \item RQ1: To what extent do frontier LLMs produce consistent annotations of misogynistic content in Polish social media discourse?
   
   
    \item RQ2: To what extent does harmful language detection transfer from general and safety-oriented classification settings to the specific task of misogyny detection?
    
    \item RQ3: How effectively do different model families detect misogynistic language in Polish? 
\end{itemize}
By answering these questions, the study contributes to both the methodological and empirical understanding of hate speech detection in under-represented languages. 

This study makes four contributions. First, we introduce a new publicly available dataset of Polish social media comments annotated for misogyny, gender stereotyping, and neutral discourse. Second, we investigate the feasibility of using frontier language models as annotators for Polish misogyny detection. Third, we provide an expert-based analysis of annotation reliability and disagreement patterns. Finally, we evaluate the cross-domain generalization abilities of existing Polish hate speech detection models on a challenging dataset containing both explicit and implicit forms of misogynistic language. Together, these contributions advance the study of hate speech detection for Polish and provide new resources and benchmarks for future research on gender-based online hostility in Slavic languages.

\section{Related Work}
Research on hate speech and offensive language detection in Polish has grown considerably in recent years; however, the ecosystem of publicly available resources remains relatively limited \citep{ptaszynski2023expert,okulska2023morpho,kolos2024ban} compared to that of English and other high-resource languages \citep[cf.][]{founta2018large,ocampo2023depth,yu2022hate,mathew2021hatexplain,de2018hate,mollas2020ethos,hartvigsen2022toxigen,zhang2018conversations,salminen2018anatomy,sap2022annotators}. Early efforts focused primarily on cyberbullying, offensive language, and general harmful content rather than on specific target groups or forms of discrimination. A notable milestone was the PolEval 2019 shared task on cyberbullying and hate speech detection \citep{kobylinski2019poleval}, which provided one of the first widely used benchmarks for harmful language identification in Polish social media content. The task addressed both binary harmful-content detection and the distinction between cyberbullying and hate speech, thereby establishing a foundation for subsequent research on automated moderation in Polish.

More recently, efforts have been made to construct larger and more realistic datasets. The BAN-PL corpus introduced content removed by professional moderators from the Wykop.pl platform and represents one of the largest openly available Polish resources for harmful language research \citep{kolos2024ban}. The dataset captures a broad spectrum of offensive and harmful online content and has significantly expanded the empirical basis for training and evaluating content moderation systems in Polish. 
Despite these advances, most existing Polish datasets treat harmful language as a relatively broad phenomenon encompassing hate speech, cyberbullying, insults, profanity, or offensive communication. Consequently, they provide limited opportunities to investigate hostility directed toward specific social groups. In particular, gender-based hostility remain substantially under-represented. Existing benchmarks therefore offer limited evidence regarding how well models trained on general hate speech datasets can transfer to more specialized forms of harmful discourse. 

Misogyny detection has emerged as a distinct research area within hate speech detection, motivated by the recognition that hostility directed at women often exhibits linguistic characteristics that differ from other forms of abusive language \citep{almanea2022armis,fersini2018overview2,fersini2022semeval,mulki2021let}. Most research in this area has been driven by shared tasks competitions. These initiatives provided annotated corpora of social media posts and introduced more fine-grained taxonomies capturing different manifestations of misogynistic language, including stereotyping, objectification, dominance, and threats of violence. Subsequent studies expanded misogyny detection to multilingual settings and explored cross-lingual transfer, demonstrating both the feasibility and the difficulties of modeling gender-based hostility across languages and cultural contexts \citep{frenda2018automatic,pamungkas2020misogyny}. In contrast, Polish remains largely absent from this body of work. Although several Polish datasets include examples of offensive or hateful language, they were not specifically designed to capture misogynistic discourse. As a result, the extent to which existing Polish NLP models can recognize misogyny, particularly in its less explicit forms, remains largely unexplored. The absence of dedicated resources also limits comparative research between Polish and other European languages, where misogyny detection has become a more established topic.

\section{Methodology}

\subsection{Corpus Construction and Annotation}
The dataset was constructed to capture explicit and implicit manifestations of misogynistic discourse in contemporary Polish social media. To this end, we collected user comments published under Facebook posts related to a highly publicized media controversy involving Anna Lewandowska, a prominent Polish public figure and influencer. The selected discussion threads generated substantial public engagement and contained a wide spectrum of reactions ranging from support to overt hostility, contempt, and stereotype-based criticism directed at women. The choice of this domain was motivated by two considerations. First, discussions concerning female public figures frequently trigger gender-oriented evaluations that extend beyond criticism of specific actions and instead invoke broader assumptions regarding women's social roles, expected behavior, and perceived responsibilities. Second, such discussions provide numerous examples of implicit forms of hostility that would be unlikely to be captured by traditional keyword-based approaches to hate speech detection. Following data collection, comments were anonymized and prepared for annotation. Each comment was assigned one of three mutually exclusive labels. Ex. \ref{ex:misogyny} \textit{Misogyny} involves expressions of hostility, contempt, hatred, or deeply rooted prejudice directed at women as a social group or at an individual explicitly on the basis of her gender. Ex. \ref{ex:stereo} \textit{Stereotyping} refers to statements attributing generalized characteristics, behaviours, competencies, or social roles to women as a group; unlike misogyny, stereotype-based comments do not contain overt hostility. Ex. \ref{ex:neutral} \textit{Neutral} category comprises comments that do not contain misogynistic content, gender stereotypes, or other forms of gender-based derogation. 

\ex. \label{ex:misogyny} \textit{Jaka to jest dziwka.}

\ex. \label{ex:stereo} \textit{Wiesniara może wyjść z obory, ale obora z wiesniary nie wyjdzie!}

\ex. \label{ex:neutral} \textit{No cóż każdy żyje jak mu pasuje.}

To increase annotation reliability while reducing the costs associated with large-scale manual coding, the dataset was annotated independently by two frontier LLMs: GPT-5.5 and Grok-4.3. Both models received identical annotation guidelines and were instructed to assign one of the three predefined labels to each comment. 
Cases in which the models produced conflicting labels were subjected to expert adjudication who assigned the final label. To evaluate annotation quality, an expert independently annotated a randomly selected subset of 100 comments. These annotations were compared against the labels generated by each language model to assess the extent to which LLM-based annotation reproduces expert judgments. Inter-annotator agreement was measured using standard agreement metrics, including percentage agreement and Cohen's $\kappa$.

\subsection{Classification Setup}
Beyond construction of the dataset, our study investigates its usefulness as a benchmark for automatic content moderation systems. In particular, we examine whether existing Polish hate speech detection models and frontier LLMs can successfully generalize to a domain characterized by misogyny-focused discourse and implicit forms of hostility. 
To assess model generalization, we consider two complementary groups of systems. The first group consists of supervised Transformer-based language models: 1) two base models: Polish BERT\footnote{\url{https://huggingface.co/dkleczek/bert-base-polish-cased-v1}} and HerBERT\footnote{\url{https://huggingface.co/allegro/herbert-base-cased}}; and 2) two safety pre-trained models: Bielik Guard\footnote{\url{https://huggingface.co/speakleash/Bielik-Guard-0.5B-v1.1}} \citep{wrobel2026bielikguardefficientpolish} and HerBERT-Guard\footnote{\url{https://huggingface.co/NASK-PIB/HerBERT-PL-Guard}} \citep{plguard2025}. Models are fine-tuned on existing Polish hate speech datasets and subsequently evaluated on the proposed misogyny corpus. This setting reflects a cross-domain transfer scenario and allows us to quantify how well existing moderation systems generalize to misogyny-focused discourse.

The second group consists of two instruction-tuned Polish LLMs evaluated in a zero-shot setting with a definition provided. Unlike traditional classifiers, these models are not fine-tuned on the target dataset but receive natural language instructions describing the annotation schema and classification task. 
This setup allows us to investigate whether contemporary generative language models can leverage their broader linguistic and pragmatic knowledge to identify implicit forms of gender-based hostility that may be difficult for conventional classifiers. In the current study, we evaluate Bielik\footnote{\url{https://huggingface.co/speakleash/Bielik-7B-Instruct-v0.1-GPTQ}} and PLLuM \citep{kocon2025pllum} --- the former is accessed through the \texttt{pipeline} function from  the transformers library; the latter is accessed through an online interface\footnote{\url{https://pllum.clarin-pl.eu/pllum_8x7b}}

Model performance is assessed using accuracy and macro-averaged F1 score. 
Particular attention is devoted to error analysis, in which we examine confusion patterns between misogyny and gender stereotyping labels, as these categories often differ not in their surface lexical realization but in the presence of implicit evaluative or prejudicial content. This analysis provides insights into the linguistic and pragmatic phenomena that remain difficult for current NLP approaches. 
Finally, we compare the performance of conventional hate speech classifiers and instruction-tuned LLMs. This comparison enables us to assess whether recent advances in generative language modeling translate into improved recognition of subtle and implicit forms of misogyny. 

For cross-domain evaluation, we use BAN-PL, a Polish-language dataset introduced by \citet{kolos2024ban}, containing harmful and offensive content collected from the Polish social networking service Wykop.pl. The dataset is particularly useful for evaluating harmful content detection because it originates from real online interactions and focuses on content that was subject to platform moderation. It also allows us to test whether models trained to predict general harmful language can effectively detect misogynistic content, allowing us to assess how well general purpose harm detection models transfer to this specific form of harmful language. 

\begin{table}[h]
    \centering
    \begin{tabular}{llll} \toprule
    Data &   Train \#  & Valid. \# & Test \# \\ \midrule
     BAN-PL    & 800  & 200 & 200 \\ 
     Misogyny  & 693  & 174 & 174 \\ 
         \bottomrule
    \end{tabular}
    \caption{Sizes of train, validation and test sets for the classification study. }
    \label{tab:datasize}
\end{table}

For the in-domain study, we split the BAN-PL dataset into stratified train, validation and test sets. 
The training set is used for parameter estimation, while the validation set is used exclusively for model selection and hyperparameter decisions. 
The test set remains unseen until the final evaluation. 
We first evaluate the final model on the held-out test set to estimate in-domain generalization. We subsequently apply exactly the same frozen model to our misogyny corpus, without further fine-tuning or parameter optimization. The resulting performance measures zero-shot cross-domain generalization.
Here, we utilise 20\% of samples from our dataset as a test set. We perform also in-domain evaluation on our misogyny corpus, utilising 60\% of data for training and the remaining 20\% for validation purposes (see Table \ref{tab:datasize} for exact set sizes).

\section{Results} 

\subsection{Misogyny Dataset}
The final corpus contains 867 Polish social media comments, of which 446 (51.4\%) were classified as \textit{Neutral}, 377 (43.5\%) as \textit{Misogyny}, and 44 (5.1\%) as \textit{Gender Stereotyping}. The resulting distribution is relatively balanced between neutral and misogynistic content, while stereotype-based instances constitute a substantially smaller but conceptually important category. This distribution confirms that the selected discussion context provides a particularly dense source of gender-directed hostility: nearly half of the collected comments contain either misogynistic or stereotype-based content.

To evaluate the reliability of LLM-assisted annotation, we compared the labels independently assigned by GPT and Grok against the final corpus labels. We consider two variants of the task. First, we evaluate the original ternary classification. Second, we collapse the two non-neutral categories into a single harmful category, producing a binary distinction between neutral and misogynistic content. The binary evaluation allows us to separate the ability to recognize the presence of gender-directed harmful discourse from the more demanding task of distinguishing misogyny from gender stereotyping. 
GPT shows very high agreement with the final annotations in both evaluation settings. The relatively small difference between the binary and three-class results indicates that GPT largely preserves the classification performance when required to distinguish between misogyny and gender stereotyping. Grok achieves lower agreement with the final corpus labels, although its results remain strong. 

The comparison between the binary and three-class settings additionally provides insight into the nature of annotation disagreement. Both models improve when misogyny and gender stereotyping are merged, indicating that at least some annotation errors concern the boundary between different forms of gender-directed harmful discourse rather than the broader distinction between harmful and neutral content. 
Taken together, these results provide a positive answer to RQ1, while simultaneously revealing some differences between the two LLM annotators. 

To further assess the reliability of the LLM-generated annotations, an expert manually annotated a randomly selected subset of 100 comments without access to the models' predictions. Agreement with the expert annotations was higher for GPT, which achieved an accuracy of 0.79 and Cohen's $\kappa=0.60$, compared with 0.73 accuracy and $\kappa=0.53$ for Grok. These results indicate moderate to substantial agreement between LLM annotators and expert judgment, confirming the reliability of the dataset\footnote{The final dataset is openly available at: \url{https://doi.org/10.5281/zenodo.24247705}}.

\subsection{Classification Study}
\paragraph{Cross-domain evaluation.}
Table \ref{tab:result} presents the results of the binary classification experiments under different training and evaluation conditions. The experiments allow us to compare three scenarios: in-domain classification, direct transfer from general harmful language detection to misogyny detection, and transfer from models previously specialized for safety and hateful content classification. 

For the base Transformer models, the strongest performance is observed when training and testing are performed within the same domain. BERT trained and tested on BAN-PL achieves a F1 of 0.875, while HerBERT reaches 0.830. However, both models experience a substantial performance decrease when transferred directly from BAN-PL to the misogyny dataset. BERT drops from 0.875 to 0.568 F1, corresponding to a transfer gap of 0.307, while HerBERT drops from 0.830 to 0.507, producing an even larger gap of 0.323. The results therefore indicate that models that perform well in general Polish harmful language detection do not necessarily transfer effectively to the more specific task of misogyny detection.

This interpretation is further supported by training directly on the proposed misogyny corpus. Under this condition, BERT reaches 0.753 F1 and HerBERT 0.760, substantially exceeding their respective BAN-PL-to-misogyny transfer results. HerBERT provides the strongest in-domain result, with an accuracy of 0.764, while maintaining relatively balanced performance between neutral (F1 = 0.792) and misogynistic (F1 = 0.728) content. Hence, the relatively poor cross-domain results cannot be attributed simply to the difficulty of the target dataset: the same base architectures perform considerably better when exposed to target-specific examples during fine-tuning.

\begin{table*}[ht]
    \centering
    \begin{tabular}{cllllll} \toprule
   Model &  Train data & Test data & Accuracy & F1 & F1 neutral & F1 hate/misogyny \\ \midrule
   
   \multirow{3}{*}{BERT} & BAN-PL  & BAN-PL & 0.875 & 0.875 & 0.882  & 0.868 \\
       &  BAN-PL & Misogyny & 0.580 & 0.568 & 0.640 & 0.497 \\
       &  Misogyny & Misogyny & 0.753 & 0.753 & 0.754 &  0.751 \\  \midrule
   \multirow{3}{*}{HerBERT} & BAN-PL  & BAN-PL & 0.830 & 0.830 &  0.835  & 0.825  \\
       &  BAN-PL & Misogyny & 0.523 & 0.507 & 0.420 & 0.595 \\
       &  Misogyny & Misogyny & 0.764 & 0.760 & 0.792 &  0.728 \\   
        \midrule
         \midrule
   \multirow{2}{*}{HerBERT-Guard} & PolyGuard/ & BAN-PL & 0.750 & 0.749 &  0.734  & 0.764 \\  
   & WildGuard  & Misogyny & 0.701 & 0.686 & 0.755  & 0.618  \\
   \midrule
   \multirow{2}{*}{Bielik-Guard} &  \multirow{2}{*}{Sojka2}  & BAN-PL & 0.875 & 0.875 &  0.869  & 0.880  \\
   &   & Misogyny & 0.655 & 0.632 &  0.725 & 0.538  \\ 
       \bottomrule
    \end{tabular}
    \caption{Cross-domain binary classification results. BERT and HerBERT are fine-tuned either on BAN-PL or the proposed misogyny dataset, while HerBERT-Guard and Bielik-Guard are evaluated using their existing safety-oriented training without additional fine-tuning on either evaluation dataset.}
    \label{tab:result}
\end{table*}

An interesting intermediate pattern emerges for models explicitly developed for harmful-content and safety classification. HerBERT-Guard obtains 0.686 F1 on the misogyny dataset, compared with 0.749 on BAN-PL, corresponding to a considerably smaller performance decrease than observed for the BAN-PL-trained base models. Similarly, Bielik-Guard achieves 0.632 F1 on misogyny compared with 0.875 on BAN-PL. Importantly, neither Guard model is fine-tuned on the proposed misogyny dataset. HerBERT-Guard was developed as a Polish safety classifier using Polish safety data together with PolyGuard and WildGuard resources and explicitly includes hate among its unsafe categories, while Bielik-Guard was trained on the Polish Sójka resource and includes a dedicated hate/aggression dimension covering discrimination based on characteristics including gender. 

These results indicate that task-specific training remains the most effective strategy, as demonstrated by the in-domain results. Nevertheless, the comparatively strong zero-shot adaptation performance of HerBERT-Guard suggests that broader safety-oriented training can improve transfer to a specific form of harmful discourse.

\paragraph{Ternary misogyny classification.} 
In a more fine-grained setting, we evaluated the models on our original three-class annotation scheme, distinguishing, between neutral, misogyny, and gender stereotyping. For the supervised encoder-based models, we explore learning rates of $1\times10^{-5}$, $3\times10^{-5}$, and $5\times10^{-5}$, with training ranging from 10 to 20 epochs. Model configurations are selected according to their performance on the validation set, and Table~\ref{tab:result2} reports the corresponding test set results. 

\begin{table*}[h!]
    \centering
    \begin{tabular}{llllll} \toprule
   Model &  Accuracy & F1 & F1 neutral & F1 misogyny & F1 stereotyping \\ \midrule
   
  BERT    & 0.747  & 0.573 & 0.794 & 0.725 & 0.200 \\
  HerBERT & 0.753  & 0.619 & 0.810 & 0.731  & 0.316 \\

  HerBERT-Guard & 0.782 & 0.742 & 0.825  & 0.734 & 0.667 \\
  
  PLLuM 8x7B-2025 & 0.517 & 0.429 & 0.327 & 0.627 &  0.333\\
  Bielik-7B-v0.1-awq  & 0.431 & 0.364 & 0.575 & 0.421  & 0.097 \\
       
       \bottomrule
    \end{tabular}
    \caption{Performance of the ternary classification task using the misogyny dataset.}
    \label{tab:result2}
\end{table*}

Among the base encoder models, HerBERT achieves strong overall performance, with an accuracy of 0.753 and F1 of 0.619, compared with 0.747 and 0.573 for BERT, respectively. Both models perform relatively well for the two dominant categories but their performance decreases considerably for gender stereotyping. This difference is particularly relevant given the pronounced class imbalance in the corpus, where stereotyping constitutes only 5.1\% of instances, compared with 51.4\% neutral and 43.5\% misogynistic content. Accordingly, the considerably lower F1 relative to accuracy for these models largely reflects weaker performance on the minority stereotyping class rather than uniformly poor classification across all three categories. 

HerBERT-Guard, however, substantially improves upon this performance pattern. It achieves the highest overall accuracy (0.782) and F1 (0.742), outperforming both BERT and HerBERT. The improvement is particularly pronounced for gender stereotyping (0.667). At the same time, its performance on the two dominant categories remains strong, outperforming slightly all other models. Bielik-Guard also reaches better performance than the two based models without harmful language detection pre-training; however, the gain is much smaller than in the case of HerBERT-Guard. 

The generative models exhibit a different performance profile. PLLuM 8x7B-2025 obtains substantially lower overall accuracy (0.517) and F1 (0.429), although its F1 of 0.333 for stereotyping exceeds those of the two base encoder models. Its performance is considerably stronger for misogyny (0.627) than for neutral comments (0.327), indicating an uneven distribution of classification performance across the three categories. Bielik-7B-v0.1-awq performs weakest overall, reaching 0.431 accuracy and 0.364 F1, with an F1 of only 0.097 for stereotyping language.

This performance distribution can also be contrasted with the earlier annotation experiment involving GPT-5.5 and Grok 4.3. Although these results are not directly comparable with the classification scores in Table~\ref{tab:result2}, as they were obtained on a different evaluation subset, they provide an informative point of reference. Considering accuracy alone, the supervised encoder models achieve performance comparable to that observed for the much larger GPT-5.5 and Grok 4.3. In contrast, the two generative Polish LLMs substantially underperform both the supervised encoder models and the frontier LLMs, further indicating the effectiveness of task- and safety-specific training for this classification setting. 

The results for dedicated Polish safety classifiers provide a particularly informative comparison with recent work on Polish content moderation effectiveness. \citet{plguard2025} found that a relatively compact HerBERT-based safety classifier could outperform substantially larger generative models, including under adversarial evaluation, demonstrating that model scale alone does not determine safety-classification performance. Similarly, \citet{wrobel2026bielikguardefficientpolish} report strong performance for compact Polish safety classifiers trained across several categories of harmful content, including hate and aggression. Our experiments extend these findings by evaluating such specialized safety models on another type of hateful category detection. Here we find that HerBERT-Guard not only transfers better to binary misogyny detection than models fine-tuned on general BAN-PL harmful content, but also achieves the strongest performance in the three-class experiment, outperforming frontier LLMs such as GPT and Grok models. The result suggests that safety-oriented specialization can provide useful transferable representations for specific forms of discriminatory language even when the downstream annotation taxonomy differs from that used during safety training.


\section{Discussion and Future Work}
The results of this study point to an important limitation of treating hate speech detection as a homogeneous classification problem. While models trained to recognize hateful language in general exhibit a degree of transferability to misogyny detection, this transfer is incomplete. Furthermore, our experiments demonstrate that the magnitude of this transfer gap depends substantially on the type of prior model specialization. Models trained directly on general harmful language data exhibit the largest deterioration when transferred to misogyny detection, whereas safety-specialized models retain considerably stronger performance without target-specific fine-tuning. The observed transfer indicates that general and specific forms of harmful language share exploitable linguistic structure, while differences between generally trained and safety-specialized models further suggest that transferability depends on the breadth and granularity of harmful phenomena represented during training.


Particularly relevant fro our study is recent work on Polish and multilingual model safety assessment. \citet{krasnodkebska2026safety} identify the predominantly English-centered development of safety benchmarks and safeguards as an important limitation of contemporary multilingual safety research, emphasizing that safety mechanisms that are effective in one linguistic setting cannot be assumed to generalize uniformly across languages and cultural contexts. Similarly, \citet{dadas2025evaluating} show that evaluating language models requires going beyond generic linguistic competence toward language- and culture-specific capabilities, since seemingly strong multilingual performance may conceal deficiencies in understanding culturally situated content. Our results extend this argument from linguistic and cultural competence to detection of harmful language manifestations: even within a single language, successful performance does not necessarily transfer uniformly across different operationalizations and subtypes of harm. This observation is consistent with the broader problem of distribution shift highlighted by \citet{zarzecki2026reliability}.

The Polish setting further extends the significance of these findings. Research on harmful language detection remains strongly shaped by resources developed for English and a relatively small number of other well-resourced languages. Introducing a dataset dedicated to Polish misogyny therefore provides an opportunity to investigate whether conclusions concerning general-to-specific transfer hold in a different linguistic and cultural environment. 
This is particularly relevant because categories such as hate speech and misogyny are operationalized through datasets, and their empirical boundaries need not coincide across languages, platforms, annotation schemes, or target communities. Consequently, transferability should not be considered exclusively a property of models; it is also a property of the relationship between datasets and the phenomena they operationalize. 

This observation opens a broader research agenda. Future work could examine general-to-specific transfer across other targets of hate, construct cross-category transfer matrices, and distinguish category transfer from cross-lingual and cross-domain transfer. Such analyses would help determine which aspects of hateful language constitute generalizable signals of harmfulness and which remain specific to particular targets, communities, languages, or communicative contexts. Ultimately, this may lead to an evaluation paradigm in which the robustness of moderation systems is assessed not only by their aggregate ability to detect hate, but also by the extent to which this ability transfers across the heterogeneous forms through which hate is expressed.

\section{Conclusions}
Taken together, our findings suggest that the relationship between general and specific forms of harmful language detection deserves to be treated as a research problem in its own right. The ability to recognize hate speech in aggregate does not necessarily imply an equivalent ability to recognize each of its manifestations. The Polish misogyny dataset introduced in this study provides one test case for examining this distinction, demonstrating both shared signal between general hate speech and misogyny and the limits of transferring from the former to the latter.

The contribution of this article to the NLP literature is therefore both empirical and methodological. Empirically, we provide systematic evidence of context-dependent variation in harmful language detection. An additional contribution is the construction and release of a dedicated Polish-language dataset for misogyny detection. Beyond enabling the experiments reported here, the dataset provides a resource for studying gender-targeted harmful language in a comparatively under-resourced linguistic context. As such, it can support further research on cross-domain generalization, as well as more fine-grained evaluation of harmful language detection systems. 

Methodologically, we argue for treating such variation as an important property of model evaluation rather than as a secondary fairness diagnostic. In doing so, the study connects research on harmful language detection with broader concerns about robustness, fairness, and socially situated language use, and motivates a shift from asking whether a hate speech classification model performs well on average toward asking where, for whom, and under which linguistic and social conditions it performs well or fails.

\section*{Acknowledgements}
During the preparation of this work, the author(s) used the GPT-5.6-Sol model exclusively to rephrase and enhance the linguistic quality of this work. The tool assisted in improving coherence and readability without contributing original ideas, analysis or substantive content. After using this tool/service, the author(s) reviewed and edited the content as needed and take(s) full responsibility for the content of the published article. \\
The work reported in this paper has been supported by the National Science Centre, Poland, under grant 2025/57/N/HS1/00480.

\section*{Limitations}
The proposed dataset is derived from a single social-media context centered on a highly visible controversy involving a female public figure. Consequently, the linguistic patterns represented in the corpus may reflect partially this particular topic and platform, and should not be assumed to represent Polish misogynistic discourse in general. The dataset is also characterized by a substantial class imbalance, with gender stereotyping accounting for only 5.1\% of the instances, which limits the reliability of conclusions concerning this category. Finally, although LLM-assisted annotation was complemented by expert adjudication and independent expert validation, distinctions between misogyny, stereotyping, and neutral discourse remain partly interpretative.

\section*{Ethics Statement}
This study examines potentially offensive and discriminatory social media content and therefore requires particular attention to privacy and responsible data use. The corpus was constructed from publicly available online comments; however, identifying information is excluded from the released dataset to minimize the possibility of tracing individual comments back to their authors. Because the annotation process involves subjective judgments concerning misogyny and gender stereotyping, LLM-generated annotations were not treated as definite ground truth: disagreements between the models were subjected to expert adjudication, and annotation reliability was additionally evaluated against an independently annotated expert sample. Finally, the dataset contains potentially harmful and offensive language, and we recommend that researchers using the resource account for possible exposure to such content and avoid reproducing identifiable or unnecessarily harmful examples.

\bibliography{0slavic}

\appendix
\section{Implementation Details}
Exact system prompt used in the classification task for LLMs: \textit{Jesteś ekspertem od lingwistyki i analiz językowych. Twoim zadaniem jest identyfikacja mizoginii (wrogość, nienawiść lub pogarda wymierzone konkretnie w kobiety, często służące karaniu ich za wyłamywanie się z tradycyjnych ról) i języka stereotypizacji (przypisywanie całej grupie ludzi np. kobietom sztywnych, negatywnych cech) wobec kobiet w języku polskim na próbce komentarzy z mediów społecznościowych. Dokonaj analizy i anotacji komentarzy według wzoru: przypisuj etykietę 'Mizoginia', lub 'Stereotypizacja', lub 'Neutralny' dla każdego z komentarzy. Zwracaj tylko jedną etykietę.} And its English translation: \textit{You are an expert in linguistics and linguistic analysis. Your task is to identify misogyny (hostility, hatred, or contempt directed specifically at women, often serving to punish them for deviating from traditional gender roles) and stereotyping (attributing rigid, negative characteristics to an entire group of people, e.g., women) directed at women in the Polish language, based on a sample of social media comments. Analyze and annotate the comments according to the following classification scheme. Assign one of the following labels to each comment: 'Misogyny' 'Stereotyping' 'Neutral'. Return exactly one label per comment.}

\end{document}